\documentclass[11pt]{article}

\usepackage{acl}

\usepackage{xcolor}

\usepackage[ruled,linesnumbered]{algorithm2e}
\usepackage{times}
\usepackage{latexsym}
\usepackage{amsmath}
\usepackage{enumitem}
\usepackage{bbm}
\usepackage{tcolorbox}

\usepackage{booktabs}
\usepackage{multirow}
\usepackage{tabularx}
\usepackage{booktabs}
\usepackage{subcaption}
\usepackage{float}

\usepackage[T1]{fontenc}

\usepackage[utf8]{inputenc}

\usepackage{microtype}

\usepackage{inconsolata}

\usepackage{graphicx}
\usepackage{amssymb}

\usepackage{tikz}
\usetikzlibrary{shapes.geometric, arrows.meta, positioning, calc}

\title{Scoring Edit Impact in Grammatical Error Correction via
Embedded Association Graphs}

\author{
  Qiyuan Xiao\thanks{These authors contributed equally to this work.}, 
  Xiaoman Wang\footnotemark[1],
  Yunshi Lan\thanks{Corresponding author.} \\
  School of Data Science and Engineering\\
  East China Normal University\\
  \{qyxiao, xmwang\}@stu.ecnu.edu.cn, yslan@dase.ecnu.edu.cn
}

\begin{document}
\maketitle
\begin{abstract}

A Grammatical Error Correction (GEC) system produces a sequence of edits to correct an erroneous sentence. 
The quality of these edits is typically evaluated against human annotations. 
However, a sentence may admit multiple valid corrections, and existing evaluation settings do not fully accommodate diverse application scenarios. 
Recent meta-evaluation approaches rely on human judgments across multiple references, but they are difficult to scale to large datasets. In this paper, we propose a new task, \textbf{Scoring Edit Impact in GEC}, which aims to automatically estimate the importance of edits produced by a GEC system. 
To address this task, we introduce a scoring framework based on an embedded association graph. 
The graph captures latent dependencies among edits and syntactically related edits, grouping them into coherent groups. 
We then perform perplexity-based scoring to estimate each edit's contribution to sentence fluency.
Experiments across $4$ GEC datasets, $4$ languages, and $4$ GEC systems demonstrate that our method consistently outperforms a range of baselines. 
Further analysis shows that the embedded association graph effectively captures cross-linguistic structural dependencies among edits.

\end{abstract}

\section{Introduction}
\label{sec:intro}

Grammatical Error Correction (GEC), which requires a system to automatically detect and correct grammatical errors in text, has made remarkable progress in recent years.
Early approaches adopted either sequence-to-sequence (Seq2seq)~\cite{DBLP:conf/acl/RotheMMKS20} or sequence-to-edit (Seq2edit)~\cite{DBLP:conf/bea/OmelianchukACS20} paradigms to address the task.
Recent studies have shifted towards developing LLM-based GEC systems~\cite{DBLP:journals/eswa/FangZWJZZLHC25}, which leverage the strong linguistic priors and inherent language understanding capabilities of large language models.

\begin{figure}[t!]
    \centering
    \includegraphics[width=1\linewidth]{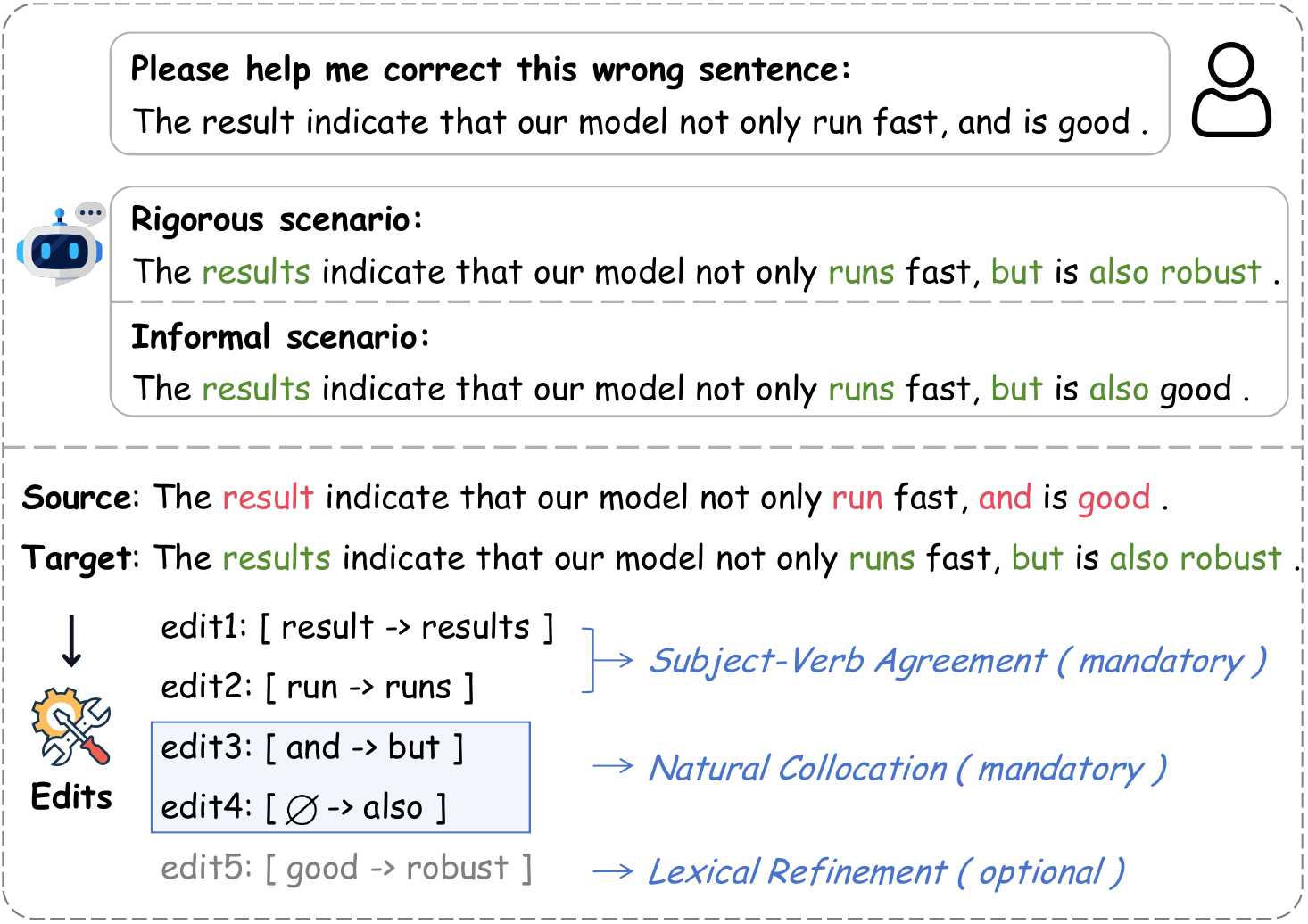}
    \caption{An example GEC instance illustrating the task of identifying edit impact. The source text indicates an erroneous text, and the target text indicates the corrected text generated by a GEC system.
    We list the corresponding edits associated with the human judgments according to their importance.}
    \label{fig:introduction example}
\end{figure}

The above GEC systems have achieved impressive performance on GEC benchmarks such as NLPCC18~\cite{DBLP:conf/nlpcc/ZhaoJS018}, FlaCGEC~\cite{DBLP:conf/cikm/DuZTWWLL23}, and CoNLL14~\cite{DBLP:conf/naacl/0004LBLZLHZ22} as measured by edit-based GEC metrics~\cite{gong:2022emnlp,DBLP:journals/tacl/KobayashiMK24}, where the quality of the edits is compared with human annotations (refer to the edits in Figure~\ref{fig:introduction example}).
However, optimal corrections for a single text may vary across different scenarios.
For example, in Figure~\ref{fig:introduction example}, the lexical refinement \textit{good} $\rightarrow$ \textit{robust} is mandatory in rigorous scenarios but optional in informal settings.
Recently, \citeauthor{DBLP:conf/naacl/0004LBLZLHZ22} (\citeyear{DBLP:conf/naacl/0004LBLZLHZ22}) proposed a multi-reference dataset that contains multiple acceptable corrections for each incorrect sentence. 
\citeauthor{DBLP:journals/tacl/KobayashiMK24} (\citeyear{DBLP:journals/tacl/KobayashiMK24}) proposed a meta-evaluation method that incorporates human rating of these multi-references.

Nevertheless, existing settings rely heavily on human annotations and cannot be easily scaled to large multilingual datasets.
Preliminary efforts~\citep{maeda-etal-2022-impara,sorokin-2022-improved} identified edit impact, which is derived from automatic evaluation metrics, to better align the effect of a single edit with human judgment.
Moving beyond identifying the impact of a single edit as a latent factor for evaluation, in this paper, we formulate the task of scoring edit impact in GEC for the first time. This task aims to rank the edits generated by a GEC system according to their importance. We formally define the task and its evaluation metrics in Section~\ref{sec:problem}.
As illustrated in Figure~\ref{fig:introduction example}, among the five edits detected by a GEC system, four are high-significance grammatical errors. 
Other edits may be plausible yet non‑mandatory.
Accordingly, we design an algorithm to automatically assign a priority score to each edit.

However, automatically scoring the impact of edits remains challenging. As illustrated in Figure~\ref{fig:introduction example}, some edits may possess comparable importance within a context; for instance, ``\textit{result}'' and ``\textit{run}'' both resolve subject-verb agreement errors. Furthermore, certain edits should be assigned equal priority due to their syntactic dependencies, such as ``\textit{not only ... but also}''.
The naive LLM-based annotation approach is inapplicable for this scoring problem due to its high computational cost and sensitivity to prompt variations\footnote{We display an experiment in Appendix~\ref{app:llm_experiment} to demonstrate the inapplicability of LLM-based annotation for edit scoring.}.
To address these issues, we propose a scoring method utilising embedded association graphs, as detailed in Section~\ref{sec:method}.
Our approach first identifies latent associations by constructing an embedded association graph, which is then used to merge syntactically dependent edits.
Subsequently, we perform perplexity-based edit scoring to assign importance scores based on each edit's contribution to text fluency.
The extensive experiments demonstrate that our method consistently outperforms a set of baselines on $4$ GEC datasets in $4$ languages under $4$ GEC systems.

In summary, our study makes the following contributions:
\begin{itemize}
    \item We formally define the task of \textbf{scoring edit impact in GEC}, which aims to estimate the relative importance of system-generated edits. This task can be integrated into writing assistants and tutoring systems to control the rigour level in GEC corrections.
    \item We propose a novel framework based on an \textbf{embedded association graph} to model latent dependencies among edits. The proposed method effectively captures edit associations and estimates their relative significance.
    \item We conduct extensive experiments across four languages and four GEC systems, demonstrating that our method consistently outperforms various baselines. Further analyses validate that the embedded association graph captures meaningful linguistic structures across different languages.
\end{itemize}
\section{Related Work}
\label{sec:related}

\subsection{GEC Evaluation}

Traditional GEC metrics, such as the MaxMatch (M2) scorer~\cite{DBLP:conf/naacl/DahlmeierN12} and ERRANT~\cite{DBLP:conf/acl/BryantFB17}, evaluate corrections by comparing token-level edits (insertions, deletions, and substitutions) with human annotations. 
Extensions such as GLEU~\cite{DBLP:conf/acl/NapolesSPT15} and ChERRANT~\cite{DBLP:conf/coling/GuHZQP25} further generalise edit-based evaluation to multilingual settings. Beyond reference-based evaluation, recent work has explored reference-free approaches that estimate correction quality directly. Early methods rely on fluency signals from n-gram models~\cite{DBLP:conf/aaai/ChollampattN18} or BERT-based representations~\cite{DBLP:conf/acl/KanekoMKSI20}, while more recent studies employ LLM-based evaluators that emphasise global fluency rather than strict edit matching~\cite{DBLP:conf/bea/KobayashiMK24}, including specialised judges such as JELV~\cite{zhan2025jelv} for validity assessment. 
A closely relevant study is IMPARA~\cite{maeda-etal-2022-impara}, which measures semantic shift in embedding space when a single edit is removed to estimate its ``impact''. While IMPARA applies the edit impact to sentence-level evaluation for a GEC system, our work has a different applied scenario: scoring the relative importance of edits as a post-processing step. Assuming the generated edits are reasonable, our goal is to estimate their relative importance, improving interpretability and enabling finer-grained control in downstream applications such as writing assistants.

\subsection{Edit Modelling for GEC}
Recent GEC research has shifted towards edit-based modelling. Seq2edit frameworks, such as LaserTagger~\cite{DBLP:conf/emnlp/MalmiKRMS19} and GECToR \cite{DBLP:conf/bea/OmelianchukACS20}, treat correction as a sequence tagging task to enhance efficiency. While effective, these methods often operate under a conditional independence assumption, struggling to coordinate discontinuous or multi-token corrections. To bridge this gap, linguistic-enhanced models have integrated dependency structures SynGEC~\cite{DBLP:conf/emnlp/0004ZLBLZ22}, constituent trees CSynGEC~\cite{DBLP:journals/corr/abs-2211-08158}, or construction grammar CxGGEC~\cite{DBLP:conf/acl/CaoWXWC25}. However, these approaches rely on syntactic formalisms or predefined inventories, which may fail to capture latent correction patterns in highly ungrammatical inputs. 
For instance, \citet{sorokin-2022-improved} proposed a two-stage framework to rank and filter erroneous edits. In contrast, our work shifts the focus to edit impact, aiming to distinguish between mandatory grammatical corrections and optional stylistic refinements. Unlike previous reranking methods that primarily handle edit conflicts, our approach explicitly mines latent associations via an embedded association graph. This allows us to preserve the structural integrity of linguistically coupled edits while accurately identifying their relative significance across diverse languages.

\section{Problem Statement}

\label{sec:problem}

\begin{figure*}[t]
  \centering
  \includegraphics[width=0.92\textwidth]{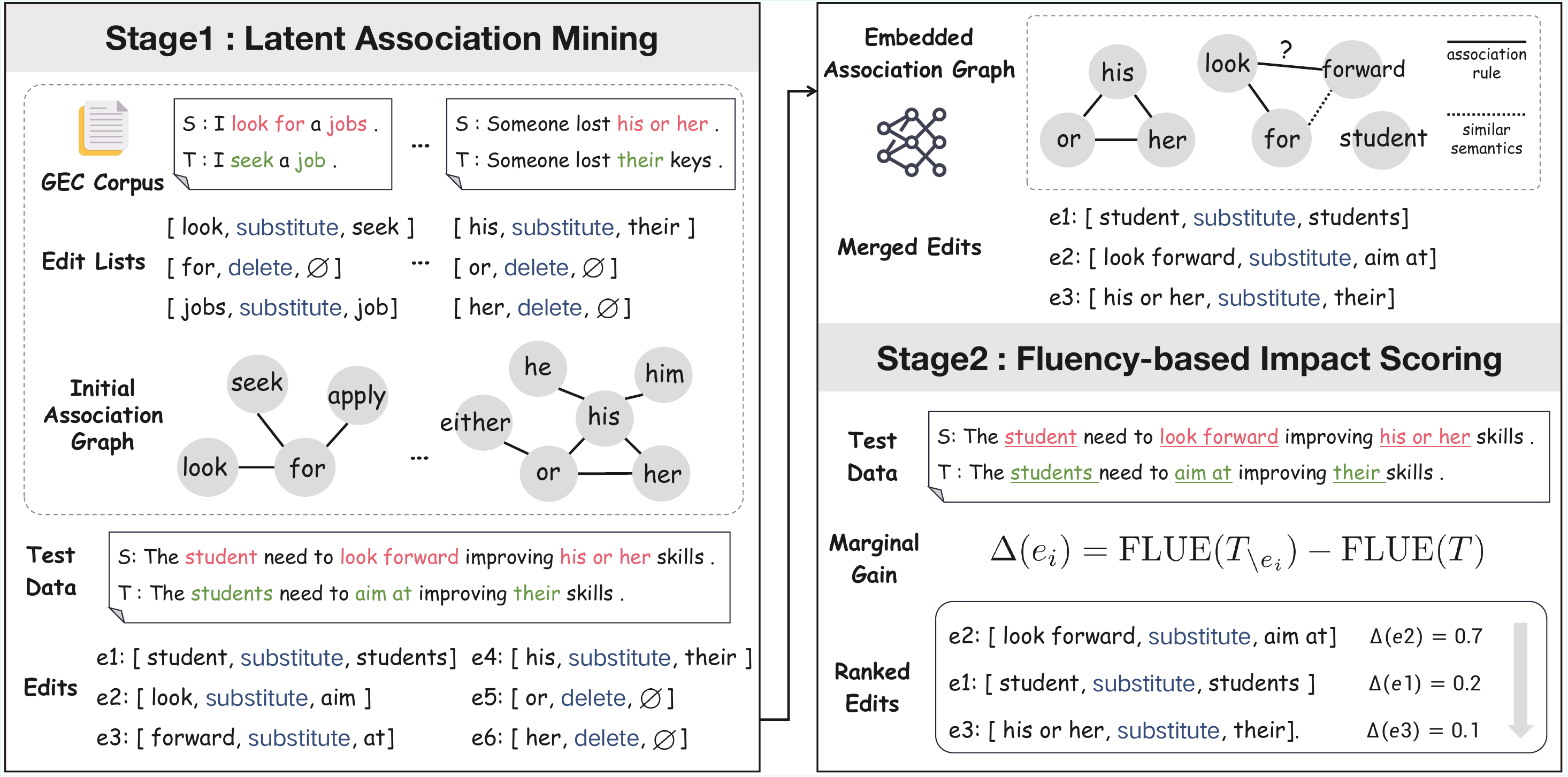}
  \caption{Overall architecture of the proposed method. Stage 1 extracts latent associations between edits, while stage 2 ranks the merged edits based on their contribution to sentence fluency reduction.}
  \label{fig:workflow}
\end{figure*}

\paragraph{Task Definition.} Grammatical error correction (GEC) aims to transform an erroneous source sentence into a grammatically correct target sentence while preserving the intended semantics. Formally, the process can be defined as follows:
\begin{align*}
    T = \text{GEC}(S),
\end{align*}
where $S$ and $T$ are source and target texts, respectively.
By mapping $S$ to $T$, we can obtain a set of individual atomic edits, denoted as $\mathcal{E}=\{e_1, e_2, ..., e_k\}$.
Each edit is represented as a triplet consisting of the source text, the edit operation (i.e., \texttt{delete}, \texttt{insert}, or \texttt{substitute}), and the target text, such as (\textit{student}, \textit{\texttt{substitute}}, \textit{students}) in Figure~\ref{fig:workflow}.

We now introduce the task of \textbf{Scoring Edit Impact in GEC}, which aims to estimate the relative importance of semantically coherent edit groups for revising an erroneous sentence. 
Given a set of edits $\mathcal{E}$, the system assigns a significance score to each edit $e \in \mathcal{E}$ and outputs an ordered list of edits according to their relative impact score:


\begin{align*}
    e_{(1)}, e_{(2)}, ..., e_{(k)} = \text{Scorer}(\mathcal{E}),
\end{align*}
where $e_{(1)}, e_{(2)}, ..., e_{(k)}$ is the list of edits in descending order by their importance and $\text{Scorer}(\cdot)$ is a model that automatically orders the set of edits according to this criterion.




The key challenges of this task are \textit{twofold}: 
\begin{enumerate}
    \item The linguistic significance of each edit should be precisely quantified  within the sentence context, which measures the fluency and completeness of the resulting sentence.
    For example, ``\textit{student}'' and ``\textit{look forward}'' in Figure~\ref{fig:workflow} belong to different error categories but both significantly impact sentence fluency.
    \item The latent associations between discontinuous atomic edits must be accurately identified and merged so that semantically or syntactically related edits are not evaluated independently. For instance, ``\textit{look forward}'' in Figure~\ref{fig:workflow} is a natural collocation. Thus, the two edits ``\textit{look}'' and ``\textit{forward}'' should be merged and assigned a shared importance score.
\end{enumerate}


\paragraph{Evaluation Metrics.}

To assess the quality of the predicted ranking $e_{(1)}, e_{(2)}, ..., e_{(k)}$, we deviate from the conventional GEC evaluation paradigm, which relies on ground-truth annotations from human experts. Instead, we adopt an automatic reference-free evaluation strategy~\cite{DBLP:conf/bea/KobayashiMK24}. This choice is motivated by two factors: First, the mapping from $S$ to $T$ varies across different GEC models, making it challenging to establish a unique gold standard for diverse model outputs.
Second, scoring edits is inherently challenging when exhibiting subtle differences and complex latent associations, necessitating a more flexible approach.

Inspired by a series of prior GEC studies on grammatical error explanation~\cite{DBLP:conf/bea/KobayashiMK24,kaneko:coling2024,li:acl2025}, we classify edits into two categories: \textsc{Corrected} (\textsc{Cor}) and \textsc{Reasonable} (\textsc{Rea}).
\textsc{Cor} edits comprise essential corrections required to eliminate grammatical errors and should thus be top-ranked.
Conversely, \textsc{Rea} edits are considered lower-priority, as they primarily serve to refine the sentence for enhanced fluency. 
We employ advanced LLMs as the annotators with quality control and prompt them to label each edit $e \in \mathcal{E}$.

Let $c_{(i)} \in \{\textsc{Cor}, \textsc{Rea}\}$ denote the label of $i$-th edit in the rank, where $N_{\textsc{Cor}} + N_{\textsc{Rea}} = k$ holds.
\begin{align*}
    & e_{(i)} \in \{e_{(1)}, ..., e_{(N_{\textsc{Cor}})}\} \quad \text{where} \quad c_{(i)} = \textsc{Cor}, \\
    & e_{(i)} \in \{e_{(N_{\textsc{Cor}}+1)}, ..., e_{(k)}\} \quad \text{where} \quad c_{(i)} = \textsc{Rea}.
\end{align*}
Then we can measure the accuracy of the ranked edits by the two defined metrics.
\begin{itemize}[leftmargin=*]
    \item $\mathbf{S_{\text{bound}}}$ (Boundary Score): This metric measures the quality of the edit rank by assuming an ideal boundary.
    We calculate the ratio of matched edits as the sum of \textsc{Rea} edits appearing above the boundary and \textsc{Cor} falling below it:    
    {
    \begin{align*}
       & \mu = \sum_{i=1}^{N_{\text{\textsc{Cor}}}} \mathbb{I}(c_{(i)} = \text{\textsc{Rea}}) + \sum_{i=N_{c}+1}^{m} \mathbb{I}(c_{(i)} = \text{\textsc{Cor}}), \\
       & S_{bound} = 1 - \frac{\mu}{k},
    \end{align*}    
    }
    where $S_{bound}$ is normalised by the total number of edits.
    $S_{bound} \in [0, 1]$ and the larger $S_{bound}$ is, the better the predicted order is.
    \item $\mathbf{S_{\text{rank}}}$ (Ranking Score): This metric penalises pairwise inversions to strictly enforce the priority $\textsc{Cor} \succ \textsc{Rea}$. 
    An inversion occurs when an annotated \textsc{Rea} edit is ranked higher than an annotated \textsc{Cor} edit. 
    The count of inversions $k$ is recorded:
    \begin{align*}
        & \sigma  = \sum_{i=1}^{k} \sum_{j=i+1}^{k} \mathbb{I}(c_{(i)} = \text{\textsc{Rea}} \land c_{(j)} = \text{\textsc{Cor}}), \\
        & S_{rank} = 1 - \frac{\sigma}{N_{\text{\textsc{Cor}}} \times N_{\text{\textsc{Rea}}} + \epsilon},
    \end{align*}
    where the ranking score is normalised by the maximum number of possible inversions plus a smoothing term $\epsilon$.
    $S_{rank} \in (0, 1]$. The larger $S_{rank}$ is, the better the predicted order is.
    This allows us to take into consideration the order within a partition. 
\end{itemize}



\section{Scoring Edit Impact via Embedded Association Graphs}
\label{sec:method}

In this section, we present our framework for identifying edit impact in GEC. The method consists of two stages: \textit{Latent Association Mining} and \textit{Fluency-based Impact Scoring}.
The former identifies latent dependencies between individual atomic edits using an embedded association graph. 
Subsequently, associated edits are merged, and the perplexity-based scoring module determines their priority by considering their marginal contribution to the erroneous sentence.
The overall workflow is illustrated in Figure~\ref{fig:workflow}.


\subsection{Latent Association Modelling}

Due to the absence of direct supervision signals, we propose a distant supervision framework to model edit associations, leveraging training signals mined from a large-scale corpus~\cite{wang:emnlp2018,beltagy:naacl2019}. Specifically, we first extract initial edit associations via heuristic rules from the training set, which serve as pseudo-labels for learning latent associations. To enhance generalisation, we construct an embedded association graph by incorporating the edits from the test data, thereby broadening the coverage of semantic dependencies. Finally, individual atomic edits are merged based on these predicted associations to form coherent edit groups, facilitating more robust scoring in the subsequent stage.

\paragraph{Mining initial associations.} 
Given a large-scale GEC corpus constructed by combining the training sets of four public GEC benchmarks listed in Table~\ref{tab:stats}, we first extract the edits for each pair.
Following the conventions of association rule mining, the set of edits within a single sentence is treated as a transaction, while each individual edit is viewed as an item. We then apply the Apriori algorithm~\cite{hegland2005apriori} to identify frequent edit patterns\footnote{The pseudocode of the algorithm is provided in Appendix~\ref{app:association}.}. For each pair of edits, we calculate their Jaccard similarity based on their co-occurrence frequencies in the corpus. Pairs exceeding a predefined threshold are selected as positive samples, indicating strong latent associations. These associations are represented as tuples $(w_i, w_j)$, which serve as the foundation for subsequent embedding-based modelling.
For example, we obtain tuples such as $(\textit{look}, \textit{for})$, $(\textit{his}, \textit{her})$, and $(\textit{his}, \textit{or})$ by mining from the GEC corpus as illustrated in Figure~\ref{fig:workflow}.

\paragraph{Modelling embedded association graph.} 
We represent the mined associations as an initial association graph. However, this graph is inherently sparse and limited in capturing the vast space of potential edit combinations in large-scale GEC tasks. To mitigate this, we incorporate token-level semantics into the graph representation. We hypothesise that semantic proximity implies relational similarity: if an association exists between two tokens, a similar link is likely to exist between their semantic neighbours. For example, if ``\textit{look}'' and ``\textit{for}'' are identified as a coherent pair requiring joint correction, a similar association should be inferred for the pair ``\textit{look}'' and ``\textit{forward}'' due to the high semantic proximity between ``\textit{for}'' and ``\textit{forward}'' in the embedding space.

Consequently, we represent each node in the graph using dense embeddings, enabling the model to implicitly generalise the association graph to unobserved edit pairs. Inspired by knowledge graph completion~\cite{nickel:2011icml,trouillon:icml2016}, we employ a negative sampling strategy where entities that do not co-occur in the corpus are treated as negative samples. The latent associations are then predicted via a classifier as follows:

\begin{equation}
    \mathbf{w}_i = \text{Emb}(w_i), \quad \mathbf{w}_j = \text{Emb}(w_j)
\end{equation}
\begin{equation}
    \mathbf{x}_{ij} = \mathbf{w}_i \oplus \mathbf{w}_j \oplus \cos_{ij} \oplus (\mathbf{w}_i \odot \mathbf{w}_j)
\end{equation}
\begin{equation}
    r = \sigma(\Phi(\mathbf{x}_{ij}))
\end{equation}
where $\text{Emb}(\cdot)$ denotes a pre-trained encoder.
The fused feature $\mathbf{x}_{ij}$ incorporates the concatenated embeddings, their cosine similarity, and their Hadamard product.
The classifier $\Phi(\cdot)$ then predicts the association probability $r$ using the sigmoid activation function $\sigma(\cdot)$.
The model is trained using binary cross-entropy loss. To mitigate class imbalance between positive associations and negative samples, we employ a balanced sampling strategy~\cite{DBLP:journals/nn/BudaMM18} during training. 

\paragraph{Merging the edits.}
At inference time, we construct an undirected graph where nodes represent atomic edits. An edge is established between edits $e_i$ and $e_j$ if and only if they satisfy the following criteria: 1) The predicted association probability exceeds a threshold $\tau$; 2) The distance between two edits in the sequence and dependency tree is within $\delta$. Finally, we identify the \textit{connected components} of the graph. Edits within the same component are merged, resulting in a set of unified edit groups $\tilde{\mathcal{E}} = \{e_1, ..., e_l\}$, where $l \leq k$.


\subsection{Fluency-based Impact Scoring}
Having addressed the latent associations between edits, we now rank the merged edit groups according to their relative importance. Following reference-free Natural Language Generation (NLG) evaluation paradigms~\cite{gong:2022emnlp,meister:acl2021}, we hypothesize that essential edits are crucial to the fluency and completeness of a sentence. Hence, we define the indicator of \textit{marginal fluency gain} of an edit given a sentence for correction
\begin{align}
    \Delta(e_i) & = \text{FLUE} (T_{\setminus e_i}) - \text{FLUE} (T),
    \label{eq:indicator}
\end{align}
where $\text{FLUE}(\cdot)$ denotes the fluency score (e.g. perplexity, BERTscore) of a sequence $T$ is the sentence with all edits applied, and $T_{\setminus e_i}$ represents the sequence where all identified edits are applied except for $e_i$.
The larger $\Delta(e_i)$ indicates that $e_i$ has a more significant impact on the sentence's fluency, as its omission results in a more substantial degradation in fluency score.
Hence, we can select the edit with the largest $\Delta(e_i)$ from the edit set.
\begin{align*}
    e_{(t)} & = \operatorname*{arg\,max}_{e_i \in \tilde{\mathcal{E}}} \Delta(e_i).
\end{align*}
We remove the selected edit from the edit set iteratively and repeatedly select the remaining largest edits.
As a result, we obtain a priority rank of edits: $(e_{(1)}, e_{(2)}, ..., e_{(l)})$.
The rank can be mapped back to $(e_{(1)}, e_{(2)}, ..., e_{(k)})$, where the merged edits share the same order in the rank.

\section{Experimental Setup}
\label{sec:setup}

\subsection{Datasets and Evaluation}
\label{ssec:datasets}
We evaluate our approach on GEC benchmarks in four different languages: \textbf{NLPCC18}~\cite{DBLP:conf/nlpcc/ZhaoJS018} for
Chinese, \textbf{CoNLL14}~\cite{ng-etal-2014-conll} for English, \textbf{COWSL2H}~\cite{DBLP:conf/lrec/DavidsonYMCGS20} for Spanish, and \textbf{FALKO}~\cite{hirschmann2022falko} for German. 
Since our method could rank the edits generated by any GEC system, we consider the standard annotation from the original datasets, the predicted target sentences from GPT-4o, GECToR~\cite{DBLP:conf/bea/OmelianchukACS20}, and T5-large~\cite{xue-etal-2021-mt5} as the target text.
We denote them as \textbf{Standards}, \textbf{GPT-4o}, \textbf{GECToR}, and \textbf{T5}, respectively.
To ensure sufficient complexity for scoring edit impact, we retain sentence pairs with at least three distinct edits ($|\mathcal{E}| \ge 3$). 
This provides a more granular testbed for assessing scoring rationality by comparing our results with those of alternative methods.
Table~\ref{tab:stats} summarises the dataset statistics. 

\begin{table}[h]
    \centering
    \renewcommand{\arraystretch}{1.1}
    \setlength{\tabcolsep}{4pt}
    \resizebox{\columnwidth}{!}{
        \begin{tabular}{llccc}
            \toprule
                \textbf{Dataset} & \textbf{System} & \textbf{\# Sents} & \textbf{Avg. Len} & \textbf{Avg. Edits} \\
            \midrule
                \multirow{4}{*}{\textbf{NLPCC18}(ZH)} 
                    & Standards & $462$ & $25.44$ & $3.72$ \\ 
                    & GPT-4o       & $479$ & $36.89$ & $4.19$ \\ 
                    & GECToR       & $147$ & $23.40$ & $3.34$ \\ 
                    & T5           & $66$  & $27.58$ & $3.35$ \\ 
            \midrule
                \multirow{4}{*}{\textbf{CoNLL14}(EN)} 
                    & Standards & $624$ & $28.88$ & $4.85$ \\ 
                    & GPT-4o       & $462$ & $30.65$ & $4.38$ \\ 
                    & GECToR       & $161$ & $33.17$ & $3.71$ \\ 
                    & T5           & $256$ & $31.53$ & $3.84$ \\ 
            \midrule
                \multirow{4}{*}{\textbf{COWSL2H}(ES)} 
                    & Standards & $221$ & $19.14$ & $3.93$ \\ 
                    & GPT-4o       & $279$ & $19.40$ & $4.03$ \\ 
                    & GECToR       & $90$  & $18.71$ & $3.53$ \\ 
                    & T5           & $153$ & $19.22$ & $3.80$ \\ 
            \midrule
                \multirow{4}{*}{\textbf{FALKO}(DE)} 
                    & Standards & $917$  & $21.73$ & $5.35$ \\ 
                    & GPT-4o       & $1092$ & $21.02$ & $5.72$ \\ 
                    & GECToR       & $352$  & $24.46$ & $4.22$ \\ 
                    & T5           & $741$  & $21.79$ & $4.85$ \\ 
            \bottomrule
        \end{tabular}
    }
    \caption{Statistics of GEC datasets computed over gold-standard annotations and system outputs from GPT-4o, GECToR, and T5.}
    \label{tab:stats}
\end{table}

We implement the evaluation procedure mentioned in Section~\ref{sec:problem}.
Given the edit sets obtained from source to target text mapping, which could be derived from various GEC models, we iteratively leave one edit out from the target sentence, which allows us to isolate the impact of an individual edit on the sentence integrity.
We employ Qwen3-8B with the temperature set to $0$ as the judge to categorise each edit as either \textsc{Reasonable} or \textsc{Corrected}, the effect of the annotation scheme is also qualified in Section~\ref{sec:consistency}.
Then we calculate both $\mathbf{S_{bound}}$ and $\mathbf{S_{rank}}$ as the evaluation metrics for each instance.
The final scores are averaged across all instances.

\subsection{Comparable Methods}
\label{ssec:baselines}

We compare our method against four baselines:

\begin{itemize}[leftmargin=*, noitemsep, topsep=0pt]
    \item \textbf{Ours}: We implement our adaptive error scoring method with an embedded association graph. We apply perplexity\footnote{\url{https://huggingface.co/docs/transformers/perplexity}} as the fluency score to derive the indicator in Equation~(\ref{eq:indicator}) due to its wide usage in measuring the fluency as an evaluation metric~\cite{gong:2022emnlp}.
    \item \textbf{Random}: Edits are prioritized in a random order.
    \item \textbf{Vanilla Scorer}: This method ranks edits by the perplexity increase $\Delta (e_i)$ when each edit is removed from the fully corrected sentence.
    This principle borrows the prior study on edit scoring~\cite{sorokin-2022-improved}.
    \item \textbf{Greedy Scorer}: A sequential selection strategy that iteratively picks the edit yielding the highest $\Delta (e_i)$ until all edits are ranked. 
    Unlike our method, it treats all edits as individual atomic units without latent association modelling.
    \item \textbf{Displacy Scorer}: An advanced baseline that groups edits based on spaCy dependency trees~\footnote{\url{https://spacy.io/}}. Two edits are merged if their affected tokens share a direct parent--child dependency relation, and the child's dependency label belongs to a predefined set of syntactic relations that typically indicate strong structural coupling.

\end{itemize}

\subsection{Implementation}
We employ Qwen3-Embedding-8B as the pre-trained encoder. 
The classifier $\Phi(\cdot)$ is implemented as a 3-layer residual MLP with a hidden size of 128. For edit merging, we set the confidence threshold $\tau=0.6$, sequence distance $\delta_s=8$, and dependency distance $\delta_d=12$ for all languages except German, where $\tau=0.75$ and $\delta_s=12$. These hyperparameters are determined based on language-specific syntactic characteristics and the scale of mined edit associations. 
More details are provided in Appendix~\ref{ap:implementation}.


\section{Experiments}
\label{sec:exp}

\begin{table*}[t]
\small
\centering
\begin{tabular}{l | l | c c | c c | c c | c c }
\midrule
GEC & Method & \multicolumn{2}{c}{\textbf{NLPCC}(ZH)} & \multicolumn{2}{c}{\textbf{CoNLL14}(EN)} & \multicolumn{2}{c}{\textbf{COWSL2H}(ES)} & \multicolumn{2}{c}{\textbf{FALKO}(DE)} \\ 
System &        & $\mathbf{S_{bound}}$ & $\mathbf{S_{rank}}$ & $\mathbf{S_{bound}}$ & $\mathbf{S_{rank}}$ & $\mathbf{S_{bound}}$ & $\mathbf{S_{rank}}$ & $\mathbf{S_{bound}}$ & $\mathbf{S_{rank}}$ \\
\midrule
\multirow{5}{*}{T5}
& Random          & $72.27$ & $64.62$ & $78.49$ & $73.80$ & $79.28$ & $76.07$ & $77.59$ & $72.70$ \\
& Vanilla Scorer & $77.85$ & $76.49$ & $84.03$ & $83.49$ & $89.50$ & $90.77$ & $84.52$ & $82.54$ \\
& Greedy Scorer  & $76.94$ & $75.03$ & $82.11$ & $79.97$ & $87.46$ & $87.87$ & $79.02$ & $73.47$ \\
& Displacy Scorer & $77.95$ & $75.78$ & $84.09$ & $82.31$ & $90.92$ & $91.97$ & $86.10$ & $82.82$ \\
& Ours            & $\mathbf{86.87}$ & $\mathbf{84.09}$ & $\mathbf{88.30}$ & $\mathbf{87.71}$ & $\mathbf{96.19}$ & $\mathbf{97.14}$ & $\mathbf{86.84}$ & $\mathbf{86.20}$ \\
\midrule 

\multirow{5}{*}{GECToR}
& Random          & $69.84$ & $65.99$ & $79.61$ & $77.74$ & $83.22$ & $80.09$ & $77.82$ & $73.66$  \\
& Vanilla Scorer & $84.42$ & $82.59$ & $88.24$ & $87.96$ & $88.52$ & $87.78$ & $78.97$ & $74.29$  \\
& Greedy Scorer  & $83.06$ & $82.29$ & $86.51$ & $83.75$ & $88.04$ & $86.93$ & $80.24$ & $76.08$ \\
& Displacy Scorer & $83.40$ & $82.43$ & $87.32$ & $85.56$ & $88.41$ & $87.87$ & $79.23$ & $72.63$  \\
& Ours            & $\mathbf{85.44}$ & $\mathbf{84.67}$ & $\mathbf{90.45}$ & $\mathbf{89.11}$ & $\mathbf{92.15}$ & $\mathbf{91.67}$ & $\mathbf{85.40}$ & $\mathbf{85.10}$  \\
\midrule 

\multirow{5}{*}{GPT-4o}
& Random          & $66.45$ & $60.07$ & $71.59$ & $64.93$ & $70.37$ & $64.76$ & $70.25$ & $62.07$  \\
& Vanilla Scorer & $79.91$ & $79.86$ & $77.23$ & $75.33$ & $79.64$ & $78.37$ & $79.00$ & $76.55$  \\
& Greedy Scorer  & $75.77$ & $72.85$ & $73.11$ & $67.84$ & $76.76$ & $74.33$ & $75.09$ & $70.81$  \\
& Displacy Scorer & $78.09$ & $75.60$ & $75.23$ & $70.79$ & $79.86$ & $78.43$ & $79.43$ & $74.87$  \\
& Ours            & $\mathbf{83.50}$ & $\mathbf{82.72}$ & $\mathbf{82.34}$ & $\mathbf{80.52}$ & $\mathbf{86.55}$ & $\mathbf{85.69}$ & $\mathbf{86.60}$ & $\mathbf{86.20}$  \\
\midrule

\multirow{5}{*}{Standards}
& Random          & $66.94$ & $62.93$ & $75.52$ & $69.87$ & $71.44$ & $63.07$ & $73.51$ & $65.65$   \\
& Vanilla Scorer & $80.15$ & $80.60$ & $82.53$ & $82.82$ & $80.92$ & $80.18$ & $81.57$ & $78.98$   \\
& Greedy Scorer  & $78.61$ & $77.03$ & $77.58$ & $72.78$ & $79.93$ & $78.02$ & $76.65$ & $70.72$  \\
& Displacy Scorer & $80.31$ & $78.79$ & $78.30$ & $74.40$ & $80.46$ & $80.10$ & $82.40$ & $78.22$  \\
& Ours            & $\mathbf{87.17}$ & $\mathbf{86.91}$ & $\mathbf{86.33}$ & $\mathbf{86.17}$ & $\mathbf{88.63}$ & $\mathbf{88.10}$ & $\mathbf{87.98}$ & $\mathbf{86.83}$  \\
\midrule
\end{tabular}
\caption{Main results for scoring edit impact in grammatical error correction.}
\label{tab:main}
\end{table*}

\subsection{Main Results}
Table~\ref{tab:main} presents the main results for scoring edit impact across various setups.
We summarise three main observations:

\begin{itemize}[leftmargin=*]
    \item \textbf{Consistent superiority over all scoring baselines.}
    Our method consistently outperforms other comparable baselines across all GEC systems and datasets, indicating that latent association modelling and perplexity-based measurement are crucial for effective grammatical error editing.
    \item \textbf{Robustness across diverse GEC systems.} 
    The performance gains remain stable across various edit sources, including Seq2Seq model (T5), Seq2Edit model (GECToR), or LLMs (GPT-4o). This demonstrates that our framework is agnostic to the underlying GEC architecture and can effectively rectify or prioritise edits from various error correction paradigms.
    \item \textbf{Strong generalisation capability across multiple languages.} 
    The effectiveness of our embedded association graph is verified across four different languages. Consistent gains in both $S_{bound}$ and $S_{rank}$ show that our approach successfully captures universal patterns of edit dependencies and necessity, proving its applicability to diverse linguistic typologies and benchmarks. Multilingual case studies can be found in Appendix~\ref{sec:case}.

\end{itemize}

\subsection{Consistency of Annotation Scheme}
\label{sec:consistency}
\begin{figure*}[t]
  \centering
  \includegraphics[width=\textwidth]{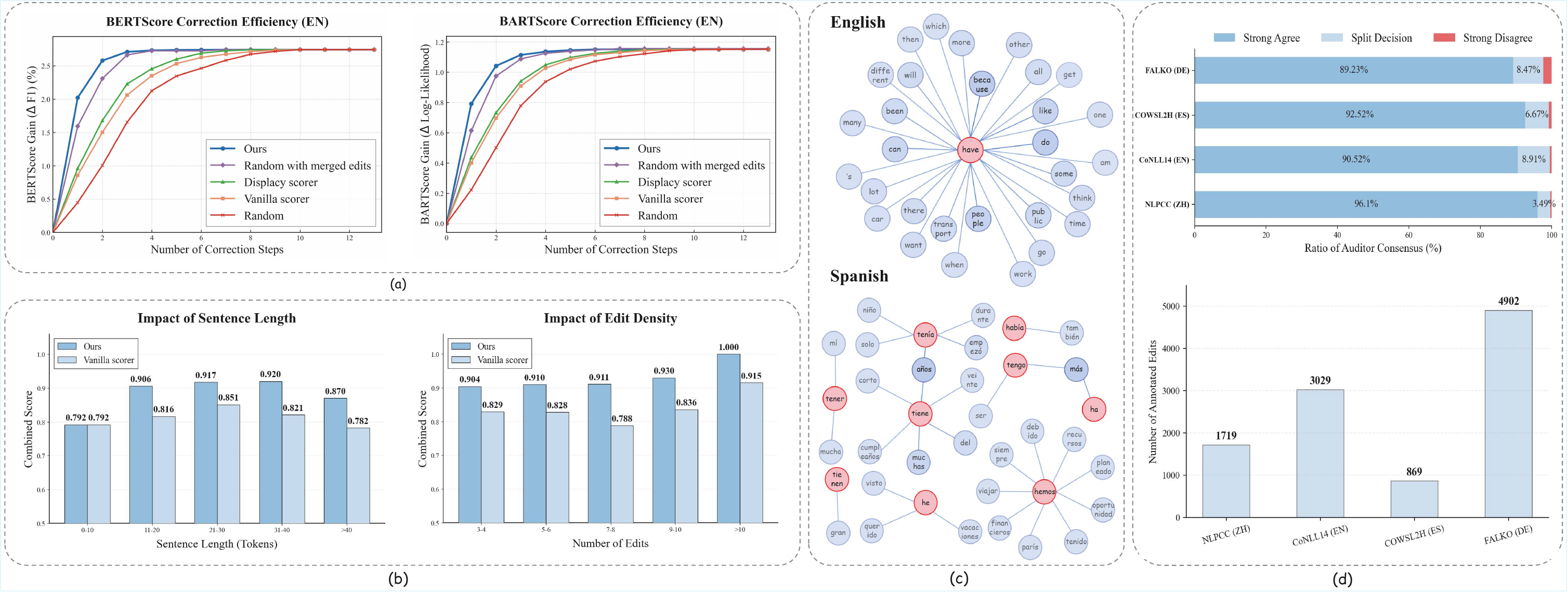}
  \caption{Experimental analysis and visualization of the proposed method. (a) The results of BERTScore and BARTScore measured by incremental gains with increasing applied edits on CoNLL14 dataset under standard. (b) The results of combined scores of methods under various sentence lengths and edit density on CoNLL14 dataset. (c) Comparison of the constructed association graphs in English and Spanish. (d) Consensus audit results of Qwen-labeled edits across multilingual datasets.}
  \label{fig:overview}
\end{figure*}

To assess the reliability of the labels generated by the primary annotator (Qwen3-8B), we conduct a consensus audit with two independent LLM auditors, DeepSeek and GPT-4o\footnote{Details of the auditing procedure are provided in Appendix~\ref{appendix:secondary auditor}.}. 
Auditor feedback is categorised into three levels: \textit{Strong Agree}, where both auditors confirm the label; \textit{Strong Disagree}, where both auditors identify issues; and \textit{Split Decision}, where the auditors disagree.
As shown in Figure~\ref{fig:overview}(d), the audit demonstrates a high level of consistency across all multilingual datasets. Most edits receive unanimous approval, while complete disagreements are rare. 
These results suggest that the LLM-based annotation provides reliable labels for evaluating the edit prioritization task. 

Furthermore, to examine alignment with human judgment, we randomly sample 100 sentences from NLPCC18 and CoNLL14, and human experts annotate the edits using the same criteria. 
The agreement between human annotations and Qwen3-8B reaches a Cohen’s Kappa~\cite{mchugh2012interrater} of 0.806 on NLPCC18 and 0.808 on CoNLL14. 
This result suggests that LLM-generated annotations can serve as a reliable proxy for human judgment in assessing edit necessity.

\subsection{Further Analysis}


\noindent \textbf{Effectiveness of our framework with increasing correction steps}.
Figure~\ref{fig:overview}(a) shows that our method achieves the largest improvements in both BERTScore~\cite{DBLP:journals/corr/abs-1904-09675} and BARTScore~\cite{DBLP:conf/nips/YuanNL21} within the fewest editing steps, outperforming the random and unmerged baselines. Even under random execution order, merged edit groups remain consistently superior, indicating that modelling latent associations between edits leads to more coherent refinements.

\noindent \textbf{Stability of our framework with alternative fluency scores}.
Table~\ref{tab:metric_comparison} further shows that the framework remains stable when the internal scoring criterion in Equation~(\ref{eq:indicator}) is replaced with alternative fluency scores such as \textbf{BERTScore}, \textbf{BARTScore}. This demonstrates the robustness of our method with respect to the choice of the internal scoring function.

\begin{table}[t]
\centering
\small
\renewcommand{\arraystretch}{1.2}
\setlength{\tabcolsep}{16pt}

\begin{tabular}{lcc}
\toprule
\textbf{Method} & $\mathbf{S_{bound}}$ & $\mathbf{S_{rank}}$ \\
\midrule
Ours + BERTScore & $86.58$ & $84.15$ \\
Vanilla + BERTScore & $75.94$ & $75.13$ \\
\midrule
Ours + BARTScore & $79.55$ & $78.11$ \\
Vanilla + BARTScore & $75.70$ & $72.04$ \\
\bottomrule
\end{tabular}
\caption{Comparison of methods with alternative fluency scores on CoNLL14 dataset and GPT-4o GEC system.}
\label{tab:metric_comparison}
\end{table}

\noindent \textbf{Effect of sentence length and edit density}.
As shown in Figure~\ref{fig:overview}(b), we evaluate the robustness of our method under varying sentence lengths and edit densities. For short sentences (fewer than $10$ tokens), both methods perform similarly, as limited structural complexity reduces the need for edit aggregation. In contrast, our approach consistently outperforms the baselines in longer sentences and higher edit-density settings, demonstrating that the latent association mining effectively captures inter-edit dependencies and enables more robust scoring in complex contexts.
This experiment highlights our potential applications in edit scoring in cross-sentence dependencies\footnote{More examples of edits involving cross-sentence dependencies can be found in Appendix~\ref{appendix:cross-sentece}.}.


\noindent \textbf{Visualization of association graph}.
\label{sec:graph_topology}
We analyze linguistic impact on graph topology in English and Spanish. With the concept ``\textit{to have}'' serving as a case study, Figure~\ref{fig:overview}(c) compares the top-$30$ associations. English exhibits a dense star topology where the lemma ``\textit{have}'' acts as a monolithic hub. In contrast, Spanish reveals a fragmented topology as semantic interactions are diluted across multiple inflexions (e.g., ``\textit{tiene}'', ``\textit{hemos}'').
This morphological dilution lowers co-occurrence-based metrics.
This implies that the embedded association graph effectively reflects linguistic structures of languages, and the merged edits could be adapted to the syntactic and morphological features of a language.
\section{Conclusion}
\label{sec:con}

In this paper, we introduced a novel task of \textbf{Scoring Edit Impact in GEC} , which is to score the edits within a GEC output for better interpretability and significance judgment.
We proposed a scoring method to solve this task, which considers the latent association between edits and fluency measurement.
This method bridges the gap between edit evaluation and modeling for GEC, providing an interpretable post-processing solution.

\section*{Limitations}
\label{sec:lim}


While our neural architecture is language-agnostic, optimal performance currently benefits from empirical tuning of inference thresholds (e.g., $\tau$ and $\delta$). Although they effectively handle major syntactic structures, empirical calibration is required to accommodate language-specific phenomena (such as German separable verbs). Future work could explore adaptive mechanisms to automate this calibration across diverse languages. 


\bibliography{custom}

\appendix
\clearpage

\section{Use of Large Language Models}
The research presented in this paper, including the core ideas, experimental design, and quantitative results, is the original work of the authors. A large language model was used as a writing assistant for tasks such as polishing prose, improving clarity, and correcting grammatical errors in the manuscript. All final content was reviewed and edited by the authors to ensure that it accurately reflects our research and contributions.

\section{Inapplicability of LLM-based Scoring Methods}
\label{app:llm_experiment}

To justify the necessity of our proposed edit scoring algorithm and address the potential question of using Large Language Models (LLMs) for this task, we conducted a systematic evaluation on three state-of-the-art LLMs: \textbf{Qwen3-8B}, \textbf{DeepSeek-V3.2}, and \textbf{GPT-4o}. 

\subsection{Experimental Setup}
We randomly sampled 50 sentence blocks from the CoNLL-14 test set, each containing at least three distinct edits. For each sentence, we prompted the LLMs to rank the edits by importance using three distinct personas:
\begin{itemize}
    \item \textbf{General Assistant:} A standard helpful assistant.
    \item \textbf{Strict Linguist:} Focused on fundamental grammatical and structural correctness.
    \item \textbf{Writing Coach:} Focused on semantic clarity and native-like fluency.
\end{itemize}
We use Kendall's Rank Correlation Coefficient ($\tau$) to measure the consistency of edit impact scoring across these three personas. A high $\tau$ (approaching 1) indicates a stable ranking, while a low $\tau$ suggests that the ranking is highly sensitive to the prompt's phrasing.

\subsection{Results and Analysis}

The quantitative results are summarised in Table~\ref{tab:llm_comparison}. Our analysis reveals several critical drawbacks of LLM-based scoring:

\begin{table}[H]
\centering
\small
\resizebox{\columnwidth}{!}{
    \begin{tabular}{llcc}
        \toprule
        \textbf{Model} & \textbf{Environment} & \textbf{$\tau$} & \textbf{Latency (s/sent)} \\
        \midrule
        Qwen3-8B       & Local                & $0.6709$          & $4.83$                      \\
        DeepSeek-V3.2  & Remote               & $0.6752$          & $5.91$                      \\
        GPT-4o         & Remote               & $0.5987$          & $7.02$                      \\
        \bottomrule
    \end{tabular}
}
\caption{Consistency and efficiency analysis of LLMs for edit scoring. $\tau$ represents the average Kendall’s Rank Correlation Coefficient across different prompts.}
\label{tab:llm_comparison}
\end{table}

\paragraph{Sensitivity to Prompting Outcomes.} As shown in Table~\ref{tab:llm_comparison}, the consistency scores across all tested models are relatively low, with none exceeding 0.68. This indicates that LLMs' judgments are significantly influenced by the assigned persona rather than the objective linguistic importance of an edit. In extreme cases, we observed negative correlations (e.g., $\tau = -0.33$), where the model completely reversed the scoring of the same edits when the prompt shifted from a ``Linguistic'' to a ``General'' persona.

\paragraph{Substantial Computational Cost.} The inference latency of LLMs is prohibitively high for GEC tasks. The fastest LLM in our test (Qwen3-8B) required 4.83 seconds per sentence, which is over 10 times slower than our proposed method (0.44s). For large-scale GEC benchmarks or real-time applications, this latency represents a major bottleneck.

In conclusion, the stochastic nature and high computational overhead of LLMs make them suboptimal for fine-grained edit scoring. Our proposed data-driven association mining approach provides a more stable, efficient, and deterministic solution.

\section{Hyperparameters and Experimental Setup}
\label{ap:implementation}

We provide the detailed hyper-parameter configurations used for Association Rule Mining (ARM), Neural Association Model ($\Phi$) training, and the inference stage, as illustrated in Table~\ref{tab:train_params} and Table~\ref{tab:infer_params}. We use a unified dependency threshold $\tau_{\text{dep}}=2$, as most core edit associations fall within two dependency hops. We also list the foundation models used. All experiments were conducted on an NVIDIA A6000 GPU.

\begin{table}[t]
\centering
\small
\begin{tabularx}{\linewidth}{@{}lX@{}}
\toprule
\multicolumn{2}{c}{\textbf{Association Rule Mining (Data Preparation)}} \\
\midrule
Min Item Frequency & 5 (3 for Spanish) \\
Min Co-occurrence & 2 \\
Min Confidence & 0.1 \\
Min Lift & 1.1 \\
Min Pair Jaccard & 0.01 \\
Similarity Filter (Word Jaccard) & 0.6 \\
\midrule
\multicolumn{2}{c}{\textbf{Model Architecture}} \\
\midrule
Semantic Encoder & Qwen3-Embedding-8B \\
Scoring Model (PPL) & Qwen3-8B \\
Source GEC Models & T5-large, GECToR, GPT-4o \\
\midrule
\multicolumn{2}{c}{\textbf{Neural Association Model ($\Phi$)}} \\
\midrule
Input Embedding Dim & 4096 \\
MRL Target Dim & 256 \\
Hidden Dimension & 128 \\
Dropout Rate & 0.4 \\
Structure & 3-layer Residual MLP \\
\midrule
\multicolumn{2}{c}{\textbf{Training Configuration}} \\
\midrule
Optimizer & AdamW \\
Learning Rate & $1 \times 10^{-4}$ \\
Batch Size & 64 \\
Training Epochs & 30 \\
LR Scheduler & ReduceLROnPlateau \\
Patience & 3 \\
Weight Decay & $1 \times 10^{-2}$ \\
\bottomrule
\end{tabularx}
\caption{Detailed Model Configurations and Training Hyperparameters.}
\label{tab:train_params}
\end{table}

\begin{table}[t]
\centering
\small
\begin{tabular}{lcccc}
\toprule
\textbf{Language} & \textbf{$\tau$} & \textbf{$\delta_{seq}$} & \textbf{$\delta_{dep}$} & \textbf{Max Neg. Ratio} \\
\midrule
English & 0.60 & 8 & 2 & 3 \\
German & 0.75 & 12 & 2 & 3 \\
Chinese & 0.60 & 8 & 2 & 3 \\
Spanish & 0.60 & 8 & 2 & 3 \\
\bottomrule
\end{tabular}
\caption{Language-specific inference thresholds. $\tau$: confidence threshold; $\delta_{seq}$: sequential distance limit; $\delta_{dep}$: syntactic dependency distance limit (consistently set to 2 hops).}
\label{tab:infer_params}
\end{table}

\begin{figure*}[h]
    \centering
    \includegraphics[width=0.98\textwidth]{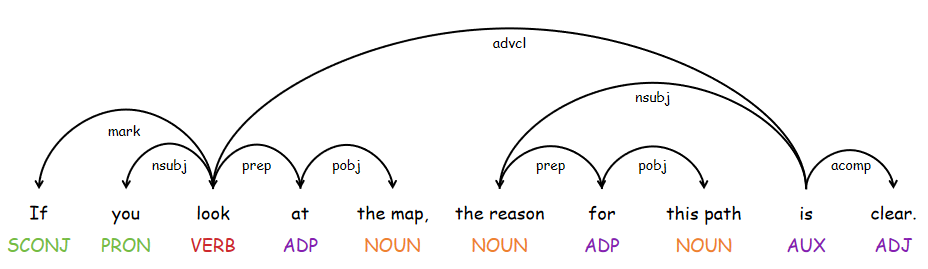}
    \caption{Dependency tree of the case.}
    \label{fig:dep_tree}
\end{figure*}

\section{Mining Edit Associations}
\label{app:association}
Algorithm~\ref{alg:association} presents the procedure for constructing the initial edit association graph from a GEC corpus. 

\begin{algorithm}[h!]
\caption{Mining Initial Edit Associations}
\label{alg:association}
\KwIn{GEC corpus $\mathcal{D} = \{(S_i, T_i)\}$, support threshold $\tau_s$, confidence threshold $\tau_c$, lift threshold $\tau_l$, similarity threshold $\tau_j$}
\KwOut{Association graph $\mathcal{G} = (\mathcal{V}, \mathcal{E})$}

$\mathcal{T} \leftarrow \emptyset$ \tcp*{transaction set}

\For{each $(S, T) \in \mathcal{D}$}{
    $\mathcal{E}_{(S,T)} \leftarrow \textsc{ExtractEdits}(S, T)$\;
    Add $\mathcal{E}_{(S,T)}$ to $\mathcal{T}$\;
}

$\mathcal{P} \leftarrow \textsc{Apriori}(\mathcal{T}, \tau_s)$ \tcp*{frequent edit pairs}

$\mathcal{A} \leftarrow \emptyset$\;

\For{each pair $(w_i, w_j) \in \mathcal{P}$}{
    $\mathcal{C}_i \leftarrow \{t \in \mathcal{T} \mid w_i \in t\}$\;
    $\mathcal{C}_j \leftarrow \{t \in \mathcal{T} \mid w_j \in t\}$\;

    Compute $J(w_i, w_j) = \frac{|\mathcal{C}_i \cap \mathcal{C}_j|}{|\mathcal{C}_i \cup \mathcal{C}_j|}$\;

    Compute $Conf(w_i, w_j) = \max\left( \frac{|\mathcal{C}_i \cap \mathcal{C}_j|}{|\mathcal{C}_i|}, \frac{|\mathcal{C}_i \cap \mathcal{C}_j|}{|\mathcal{C}_j|} \right)$;

    Compute $Lift(w_i, w_j) = \frac{|\mathcal{C}_i \cap \mathcal{C}_j| \times |\mathcal{T}|}{|\mathcal{C}_i| \times |\mathcal{C}_j|}$;

    \If{$J(w_i, w_j) > \tau_j$ \textbf{and} $Conf(w_i, w_j) > \tau_c$ \textbf{and} $Lift(w_i, w_j) > \tau_l$}{
        $\mathcal{A} \leftarrow \mathcal{A} \cup \{(w_i, w_j)\}$\;
    }
}

$\mathcal{V} \leftarrow$ set of all edits in $\mathcal{A}$\;
$\mathcal{E} \leftarrow \mathcal{A}$\;

\Return{$\mathcal{G} = (\mathcal{V}, \mathcal{E})$}
\end{algorithm}

\section{Case Study}
\label{sec:case}

\subsection{Necessity of Merging}

\label{ssec:case_merging}

We highlight the necessity of merging using a \textbf{German separable verb} case:

\begin{flushleft}
\small
\noindent\rule{1\linewidth}{1pt} \\
\textbf{  Source text}: Er \textit{fang} die \textit{Arbeit} am . \\
\textbf{  Target text}: Er \textit{fängt} die \textit{Arbeit} an . ( He begins to work . )
\noindent\rule{1\linewidth}{1pt}
\end{flushleft}

\begin{table}[H]
\centering
\small
\setlength{\tabcolsep}{4pt}
\begin{tabular}{llcc}
\toprule
\textbf{State} & \textbf{Sentence Content} & \textbf{PPL} ($\downarrow$) & \textbf{$\Delta$PPL} \\
\midrule
Source ($S$) & Er \textbf{fang} die Arbeit \textbf{am} . & 1312.5 & - \\
$S \leftarrow e_{stem}$ & ... \textbf{fängt} ... \textbf{am} . & 155.1 & +1157.4 \\
$S \leftarrow e_{prefix}$ & ... \textbf{fang} ... \textbf{an} . & \textbf{1736.0} & \textbf{-423.5} \\
Target ($T$) & ... \textbf{fängt} ... \textbf{an} . & \textbf{112.8} & +1199.7 \\
\bottomrule
\end{tabular}
\caption{A German case of separable verb}
\label{tab:case_german}
\end{table}

As shown in Table~\ref{tab:case_german}, treating edits in isolation creates a ``fluency trap''. Correcting only the prefix ($e_{pre}$: \textit{am$\to$an}) while leaving the stem erroneous (\textit{fang}) causes PPL to spike from 1312.5 to 1736.0, yielding a negative importance score (-423.5). This valid correction would thus be wrongly penalised by traditional scoring.

By merging them into $G=\{e_{stem}, e_{pre}\}$, our system evaluates their collective impact. The combined correction reduces PPL to 112.8, unlocking a massive importance score ($\Delta$PPL $\approx$ 1200). This confirms that merging is a prerequisite for accurately estimating edit value in discontinuous dependencies.

\subsection{Validity of Scoring}

A key capability of our reference-less framework is distinguishing between necessary grammatical corrections ($\textsc{Cor}$) and optional stylistic optimisations ($\textsc{Rea}$). Table~\ref{tab:case_english} illustrates this with an English example containing a tense error ($e_{tense}$: \textit{finish$\to$finished}) and a lexical choice ($e_{lex}$: \textit{task$\to$homework}).

\begin{table}[H]
\centering
\small
\setlength{\tabcolsep}{3pt}
\begin{tabular}{llcc}
\toprule
\textbf{Edit Type} & \textbf{Hypothesis Sentence} & \textbf{PPL} & \textbf{$\Delta$PPL} \\
\midrule
Source & I have \textbf{finish} my \textbf{task}. & 1391.3 & - \\
$S \leftarrow e_{lex}$ & ... \textbf{finish} ... \textbf{homework}. & 1165.1 & +226.2 \\
$S \leftarrow e_{tense}$ & ... \textbf{finished} ... \textbf{task}. & \textbf{224.9} & \textbf{+1166.4} \\
Target & ... \textbf{finished} ... \textbf{homework}. & 180.5 & +1210.8 \\
\bottomrule
\end{tabular}
\caption{Scoring analysis on English edits}
\label{tab:case_english}
\end{table}

While both edits improve sentence fluency, the grammar error correction ($e_{tense}$) yields a massive PPL drop ($\Delta \approx 1166$), whereas the stylistic swap ($e_{lex}$) provides only a marginal gain ($\Delta \approx 226$). This quantitative gap ensures that $e_{tense}$ is ranked above $e_{lex}$, directly maximizing the \textbf{Ranking Score ($S_{rank}$)}. Moreover, the distinct PPL impact facilitates the identification of the boundary between essential and optional edits, supporting a robust \textbf{Boundary Score ($S_{bound}$)}.

\subsection{Role of Syntactic Filtering}

Statistical methods like ARM often struggle with ``pseudo-associations''—pairs that frequently co-occur in the corpus but are structurally unrelated in a specific context. We illustrate this with the sentence:

\begin{center}
\small
\noindent\rule{1\linewidth}{0.5pt} \\
\vspace{2pt}
\textit{If you \textbf{look} at the map, the reason \textbf{for} this path is clear.}
\vspace{2pt}
\noindent\rule{1\linewidth}{0.5pt}
\end{center}

As shown in the dependency tree (Figure~\ref{fig:dep_tree}), \textit{look} serves as the predicate of the adverbial clause, while \textit{for} acts as a prepositional modifier in the main clause. Despite the high statistical correlation of the phrasal verb ``look for'' in our training data, these two tokens are structurally distant in this specific sentence.

Our ARM module initially assigned a high association score to this pair. However, the dependency check reveals a shortest path distance of 3 hops (\textit{look} $\leftrightarrow$ \textit{is} $\leftrightarrow$ \textit{ the reason} $\leftrightarrow$ \textit{for}), exceeding our threshold of $\delta_{dep}=2$. Consequently, the system rejects the merge. This demonstrates that the syntactic constraint acts as a crucial filter, preventing over-merging and preserving the precision of edit boundaries.

\subsection{Cross-Sentence Dependencies}
\label{appendix:cross-sentece}



We present a case study to illustrate the effectiveness of our edit scoring framework in handling cross-sentence dependencies. As shown in Table~\ref{tab:case_study_example}, our method successfully captures inter-edit associations across sentences, producing more coherent corrections than the baseline.


\begin{table*}[t]
\centering
\begin{minipage}{\linewidth}
\textbf{Original Sentence:} Social media sites are at their peaks at the moment, however, I believe that when people realize that such sites are not making them better persons in real-life, they will not be indulged in it as much as they do now.\\[0.3em]
\textbf{Target Sentence:} Social media sites are at their peak at the moment. However, I believe that when people realize that such sites are not making them better people in real life, they will not indulge in them as much as they do now.
\end{minipage}

\medskip
\resizebox{\linewidth}{!}{
\begin{tabular}{lcccccl}
\toprule
\textbf{Rank} & \textbf{Label} & \textbf{Delta F1} & \textbf{Span} & \textbf{Original Token} & \textbf{Correction} & \textbf{Edit Description} \\
\midrule
1 & corrected & 0.0157 & [6:7] & peaks & peak & Noun number correction \\
2 & corrected & 0.0071 & [36:38],[39:40] & be indulged in it & indulge in them & Verb form + pronoun agreement \\
3 & corrected & 0.0046 & [30:31] & real-life &  & Hyphen removal \\
4 & corrected & 0.0033 & [11:12] & however & However & Capitalization after period \\
5 & corrected & 0.0012 & [10:11] & , & . & Punctuation: comma to period \\
6 & reasonable & 0.0000 & [27:28] & persons & people & Lexical collocation refinement \\
\bottomrule
\end{tabular}
}

\caption{Case Study of Cross-Sentence Edit Impact Effectiveness.}
\label{tab:case_study_example}
\end{table*}

\section{LLM Prompt Templates}
\label{app:prompts}

We employ large language models (LLMs) for three prompt-based tasks in this work: (1) \textit{Consistency Auditing}, which validates the reliability of these labels using an independent expert model; (2) \textit{Necessity Labelling}, which determines whether an edit group is grammatically mandatory or stylistic; (3) \textit{Grammatical Error Correction (GEC)}, where GPT-4o serves as a zero-shot baseline.

\subsection{Prompt Design for Consistency Auditing}
\label{appendix:secondary auditor}
To enhance label reliability, we introduce a secondary auditor (DeepSeek-V3.2 and GPT-4o) to re-evaluate the necessity labels under a predefined ``tolerance'' principle. The corresponding prompt, shown in Table~\ref{tab:prompt_audit}, ensures that the final classifications remain linguistically plausible and consistent.

\begin{table}[h]
  \centering
  \small
  \setlength{\tabcolsep}{0pt}
  \begin{tabular}{p{0.98\columnwidth}}
    \toprule
    You are an experienced and tolerant language expert auditing GEC suggestions from an AI assistant. \\[0.5em]
    \textbf{Tolerance Principle:} As long as the assistant's judgment ``makes sense'' or ``is reasonable'', please provide your approval. \\[0.5em]
    \textbf{Audit Guidelines:} \\
    1. Language is flexible. If the suggestion makes the expression more idiomatic/neat (even if not a hard error), choose \textit{agree}. \\
    2. Choose \textit{disagree} only if the assistant makes a low-level error (e.g., creating a new error or ignoring obvious logical chaos). \\[0.5em]
    \textbf{Output:} Directly output \texttt{agree} or \texttt{disagree} for each task. \\
    \bottomrule
  \end{tabular}
  \caption{Prompt for cross-model label validation and consistency auditing.}
  \label{tab:prompt_audit}
\end{table}

\subsection{Prompt Design for GEC}
\label{appendix:gec_gpt4}
We use GPT-4o as an off-the-shelf, zero-shot baseline for grammatical error correction (GEC), without any task-specific fine-tuning. The model is instructed to perform sentence-level grammatical error correction under a strictly controlled instruction format. Multilingual prompt templates are designed for English, Chinese, German, and Spanish, as shown in Table~\ref{tab:gec_prompts_appendix}.

\subsection{Prompt Design for Necessity Labelling}
The primary evaluator (Qwen-3-8B) is prompted to assess the grammatical acceptability of sentences after specific edit groups are removed. Based on the resulting judgments, each edit group is labelled as either grammatically necessary or stylistic. Table~\ref{tab:prompt_multilingual} presents the multilingual prompt templates used for English, Chinese, Spanish, and German.

\begin{table*}[ht]
  \centering
  \small
  \begin{tabular}{p{0.96\textwidth}}
    \toprule
\textbf{English}\\
You are a helpful assistant assisting in evaluating English sentences. Determine if the following sentences are acceptable in general English contexts. ... [List of Sentences] ...\\
Criteria: \textit{reasonable} (acceptable); \textit{corrected} (clear errors/broken structure).\\[0.5em]

\textbf{Chinese}\\

\includegraphics[width=0.95\textwidth]{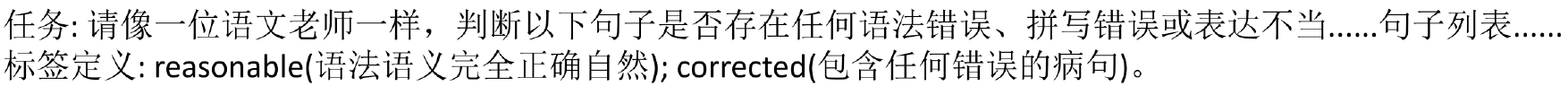}

\textbf{Spanish}\\
Actúa como un asistente útil para evaluar oraciones en español. Tu tarea es determinar si las siguientes oraciones son gramaticalmente aceptables. ... [Lista de oraciones] ...\\
\textbf{Criterios:} \textit{reasonable} (aceptable y comprensible); \textit{corrected} (errores claros o faltas de ortografía).\\[0.5em]

\textbf{German}\\
Du bist ein objektiver Korrektor für deutsche Texte. Deine Aufgabe ist es zu entscheiden, ob die folgenden Sätze grammatikalisch korrekt sind. ... [Satzliste] ...\\
Kriterien: \textit{reasonable} (grammatikalisch korrekt, Kasus/Verbformen stimmen); \textit{corrected} (enthält Grammatikfehler).\\[0.5em]

\textbf{Output Format (All Languages)}\\
Please output exactly $N$ lines. Each line must contain ONLY the word 'corrected' or 'reasonable'. Do NOT output numbering, explanations, or thinking processes.\\[0.5em]

    \bottomrule
    
  \end{tabular}
  \caption{Multilingual prompts for identifying mandatory versus stylistic edits across four languages.}
  \label{tab:prompt_multilingual}
\end{table*}

\begin{table*}[ht]
  \centering
  \small
  \begin{tabular}{p{0.96\textwidth}}
    \toprule
    \textbf{English} \\
    You are a professional grammar correction assistant. Please correct all grammatical errors and spelling mistakes in the sentence marked with the \texttt{<input>} tag. \\
    Guidelines: 1. Comprehend the sentence context; 2. Identify and correct all errors (grammar, spelling, expressions); 3. Preserve original structure/style; 4. If error-free, return unchanged; 5. Return ONLY the corrected sentence. \\
    Output Format: Please wrap the corrected sentence with the <output> tag. \\
    Please correct the following sentence: \\
    \texttt{<input> \{\} </input>}   \texttt{/no\_think} \\[0.5em]

    \textbf{Chinese}\\

    \includegraphics[width=0.95\textwidth]{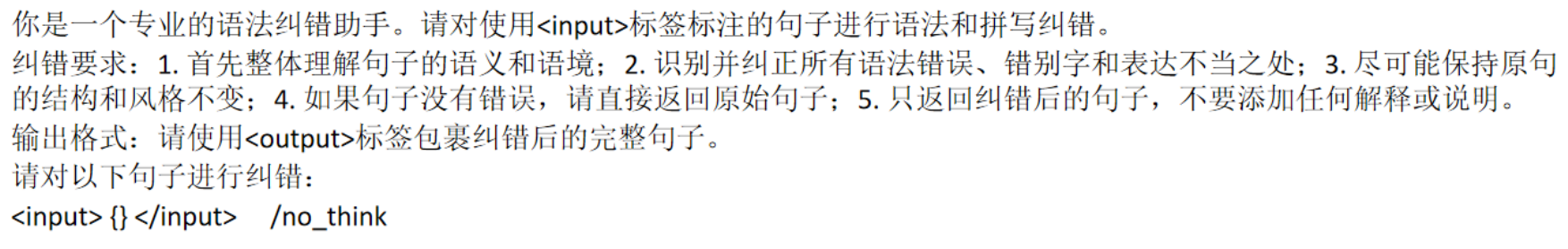}

    \textbf{German} \\
    Sie sind ein professioneller Korrekturassistent für die deutsche Sprache. Bitte korrigieren Sie alle grammatikalischen Fehler, Rechtschreibfehler und unangemessenen Ausdrücke im Satz. \\
    Richtlinien: 1. Kontext erfassen; 2. Fehler identifizieren und korrigieren; 3. Satzstruktur und Stil bewahren; 4. Bei Fehlerfreiheit unverändert zurückgeben; 5. Nur den korrigierten Satz ausgeben. \\
    Ausgabeformat: Bitte umschließen Sie den korrigierten Satz mit dem <output>-Tag. \\
    Bitte korrigieren Sie den folgenden Satz: \\
    \texttt{<input> \{\} </input>} \texttt{/no\_think} \\[0.5em]

    \textbf{Spanish} \\
    Eres un asistente profesional de corrección gramatical en español. Corrige todos los errores gramaticales, ortográficos y expresiones inadecuadas en la oración. \\
    Directrices: 1. Comprender el contexto; 2. Identificar y corregir errores; 3. Conservar estructura y estilo; 4. Si no hay errores, devolver sin cambios; 5. Devolver únicamente la oración corregida. \\
    Formato de salida: Envuelve la oración corregida con la etiqueta <output>. \\
    Por favor, corrige la siguiente oración:\\
    \texttt{<input> \{\} </input>} \texttt{/no\_think} \\    
    \bottomrule
  \end{tabular}
  \caption{Multilingual prompt templates used for LLM-based(GPT-4o) Grammatical Error Correction across four languages.}
  \label{tab:gec_prompts_appendix}
\end{table*}


\end{document}